\documentclass[10pt,a4paper]{article} 

\usepackage{times}
\usepackage{epsfig}
\usepackage{graphicx}
\usepackage{amsmath,bm}
\usepackage{amssymb}
\usepackage{algorithm}
\usepackage{subfigure}
\usepackage{algorithmic}

\usepackage[pagebackref=true,breaklinks=true,letterpaper=true,colorlinks,bookmarks=false]{hyperref}

\begin{document}

\title{Monte Carlo DropBlock  for Modelling Uncertainty  in Object Detection }

\author{Kumari Deepshikha \textsuperscript *\\
NVIDIA \\
India \\
{\tt\small deepkshikha@gmail.com}
\thanks{equal contribution, alphabetical order } 
\and
Sai Harsha Yelleni \textsuperscript *\\
IIT Hyderabad\\
Hyderabad\\
{\tt\small cs19mtech11023@iith.ac.in}
\textsuperscript *
\and
P.K. Srijith \\
IIT Hyderabad\\
Hyderabad\\
{\tt\small srijith@cse.iith.ac.in}

\and
C Krishna Mohan\\
IIT Hyderabad\\
Hyderabad\\
{\tt\small ckm@cse.iith.ac.in}
}

\date{}

\maketitle

\begin{abstract}
With the advancements made in deep learning, computer vision problems like object detection and segmentation have seen a great improvement in performance.  However, in many real-world applications such as autonomous driving vehicles, the risk associated with incorrect predictions of objects is very high. Standard deep learning models for object detection such as YOLO models are often overconfident in their predictions and do not take into account the uncertainty in predictions on out-of-distribution data.   In this work,  we propose an efficient and effective approach to model uncertainty in object detection and segmentation tasks using Monte-Carlo DropBlock (MC-DropBlock) based inference. The proposed approach applies drop-block during training time and test time on the convolutional layer of the deep learning models such as YOLO.  We show that this leads to a Bayesian convolutional neural network capable of capturing the epistemic uncertainty in the model. Additionally, we capture the aleatoric uncertainty using a Gaussian likelihood.  We demonstrate the effectiveness of the proposed approach on modeling uncertainty in object detection and segmentation tasks using out-of-distribution experiments.  Experimental results show that MC-DropBlock improves the generalization, calibration, and uncertainty modeling capabilities of YOLO models in object detection and segmentation.
\end{abstract}

\section{Introduction}

Deep learning has improved the state-of-the-art in many problems in computer vision including image classification, segmentation, and object detection \cite{he2017mask,redmon2016you,rawat2017deep}. This has helped their deployment in various real-world applications. However, the popular deep learning models suffer from several drawbacks such as lack of uncertainty modeling capability, failure in open-set conditions, and erroneous predictions with high confidence \cite{nguyen2015deep,guo2017calibration}.  This can be dangerous in high-risk applications such as autonomous driving and healthcare. Assisted driving vehicles have met with accidents due to overconfident predictions emphasizing the necessity to model uncertainty in deep learning models \cite{nhtsa17,mcallister17}. These applications have to frequently encounter out-of-distribution data, where the data come from a distribution different from the train data distribution.  For instance, in autonomous driving, it can encounter a  new type of vehicle or a random object which is not a vehicle. It is important for deep learning models to accurately estimate uncertainty in their predictions and not make overconfident predictions in these situations.


\begin{figure}%
\centering
\subfigure[Baseline YOLOv5 ]{%
\label{fig:intro}%
\includegraphics[height=1.5in]{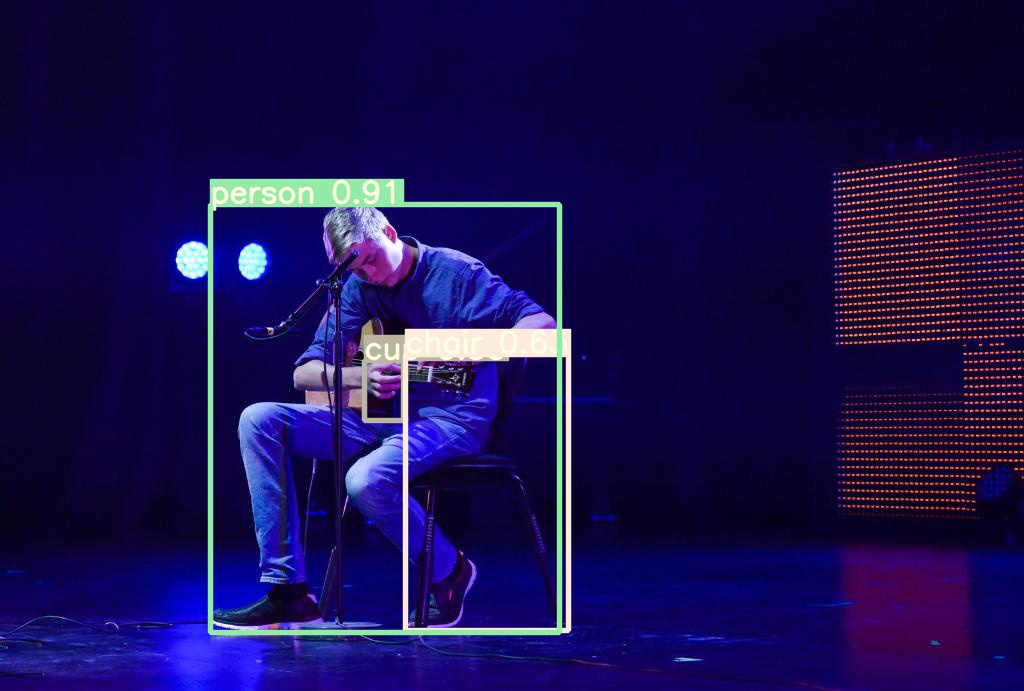}}%
\qquad
\subfigure[ MC-DropBlock YOLOv5]{%
\label{fig:intro2}%
\includegraphics[height=1.5in]{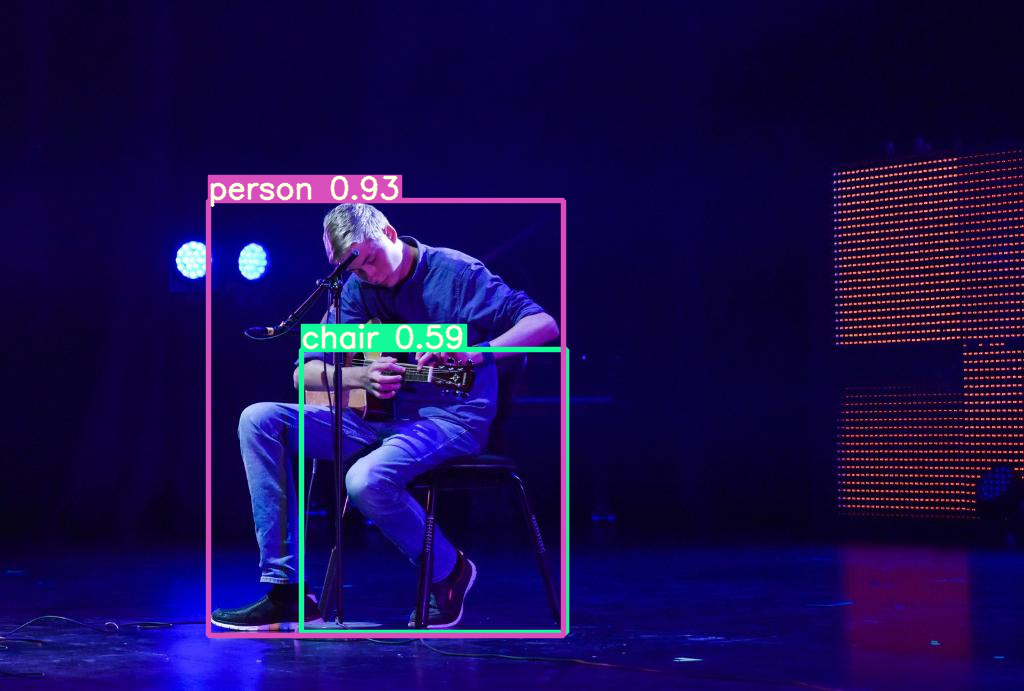}}%
\qquad
\caption{Object detection results of baseline YOLOv5 and MC-DropBlock YOLOv5, trained on COCO data and tested on an image from Open Images Dataset.}
\end{figure}

Many high-risk computer vision applications require performing tasks such as segmentation and object detection\cite{szegedy2013deep}. 
Modern deep learning algorithms used for these tasks are based on the convolutional networks\cite{lecun2010convolutional} such as Resnets\cite{szegedy2017inception}, Yolo\cite{redmon2017yolo9000}, and R-CNN\cite{ren2018object}. They make use of the spatial dependencies in the images through the convolution operation for solving the tasks more effectively.  Through multiple convolutional layers, and deeper representations they have achieved state-of-the-art results in many computer vision problems. 
Many networks have been proposed for object detection and localization  \cite{redmon2016you,liu2016ssd,bochkovskiy2020yolov4,lin2017focal,ren2016faster,bochkovskiy2020yolov4} and are extensively used in numerous applications like smart tracking, and autonomous vehicles. These  CNN-based state-of-the-art neural networks give a very good predictive performance on object detection. However, their performance is limited to in-distribution data on which training is done but does not handle uncertainties well on out-of-distribution (OOD) data. Figure~\ref{fig:intro} shows the objects detected by the YOLOv5 model trained on the COCO data and tested on an image from the Open Images data. The baseline YOLOv5 model incorrectly detects the person's hand as a cup with high confidence and the chair is also not properly localized. 

Bayesian approaches provide a principled framework to capture uncertainty in machine learning models~\cite{ghahramani2015probabilistic}. They naturally model uncertainty through posterior distribution over model parameters and Bayesian model averaging. However,  Bayesian inference techniques for neural networks~\cite{segalman2013epistemic,blundell15,kingma2015variational} are computationally costly and practically difficult to implement for many recent deep learning models for vision.  Monte Carlo Dropout~\cite{gal2016dropout} was introduced as a very efficient and effective way to implement Bayesian principles and uncertainty in deep learning models. The approach performs dropout at training time and test time, and predictions are made by taking an average over multiple dropout architectures. However, many modern deep learning models such as YOLO are based on convolution operation and consequently involve many convolutional layers. Dropout was found to be ineffective on the convolutional neural networks~\cite{wirges2019capturing} as the spatial collocation of neurons was not taken into consideration. \cite{ghiasi2018dropblock} introduced DropBlock, which overcomes this drawback by dropping neurons in a contiguous region. 

In this work, we propose Monte-Carlo DropBlock (MC-DropBlock) as an effective approach to model uncertainty in popular convolutional networks such as YOLO models used for object detection and image segmentation.  A fast and effective approach is extremely important and useful for real-time object detection in many high-risk applications. MC-DropBlock applies DropBlock at training time and at test time, and predictions are made using multiple drop block architectures.  We also derive theoretical proof showing that  MC DropBlock is equivalent to a Bayesian convolutional neural network, and consequently captures epistemic uncertainty on the out-of-distribution data. In addition, we also modeled aleatoric uncertainty by including localization uncertainty through a Gaussian likelihood. We conduct experiments using popular models such as  YOLOv4,  YOLOv5, and YOLACT on computer vision tasks such as object detection and image segmentation. The experimental results show that the proposed approach not only improves uncertainty modelling capability but also the generalization and calibration capability of the YOLO models. To the best of our knowledge,  the effect of DropBlock on the generalization capability of YOLO models was not demonstrated in any prior work and our work aims to bridge this gap as well. Moreover, we also show that MC-DropBlock can bring improvements over MC-Dropout for modeling uncertainty in image classification tasks using Resnets. The major contributions of our work can be summarized  as follows : 
\begin{itemize}
\item We propose Monte-Carlo DropBlock to model epistemic uncertainty in convolutional neural networks such as YOLO models for object detection and segmentation.
\item We theoretically show that applying MC-DropBlock is equivalent to performing variational inference on a  Bayesian convolutional neural network. 
\item We show quantitatively and qualitatively that MC-DropBlock improves generalization, calibration and uncertainty modeling capabilities of YOLO models   for object detection and segmentation.
\end{itemize}

\section{Related Work}

There are two types of uncertainty associated with model predictions: epistemic uncertainty~\cite{ren2016faster} and aleatoric uncertainty~\cite{kendall2017uncertainties}. Epistemic uncertainty describes what the model does not know due to limitations in data and ignorance about the model which has generated the data. Aleatoric uncertainty is the uncertainty arising from the natural stochasticity in  observations~\cite{costa2012towards}. It captures the inherent noise in the data and cannot be reduced by collecting more data. For Computer Vision, Bayesian Deep Learning has shown some significant performance to model uncertainties\cite{kendall2017uncertainties}. To handle epistemic uncertainty, Bayesian Neural Networks \cite{mackay1995probable} put a prior over the parameters and compute a posterior distribution over them using observed data. Monte-Carlo Dropout~\cite{gal2016dropout} successfully developed a Bayesian neural network by applying dropout at train time and test time. This allowed them to obtain model uncertainty out of existing deep learning models and made Bayesian deep learning practical and easy to apply to several deep learning models. 

Successful deep learning models for object detection, segmentation, and classification such as Resnets~\cite{szegedy2017inception} and YOLO~\cite{redmon2018yolov3}  models are based on convolutional neural networks which involve several convolutional layers. For object detection, YOLOv4~\cite{bochkovskiy2020yolov4}
is the state-of-the-art method for real-time object detection.  Though Dropout was proven to be an effective regularization method for neural networks, they are less effective for fully convolutional layers.   DropBlock~\cite{ghiasi2018dropblock} has been introduced which is an effective regularization method, especially for convolutional layers, because of their ability to drop neurons in a contiguous region and stop spatial information flow. 

There are some recent works in considering uncertainty in object detection models. Gaussian YOLOv3~\cite{choi2019gaussian} is proposed to capture aleatoric uncertainty.  The architecture of YOLOv3 \cite{redmon2018yolov3} is added with an extra layer, which puts a Gaussian distribution over each bounding box co-ordinate with variance giving the uncertainty of the co-ordinate. To accommodate the change made in the prediction head, a negative log-likelihood loss is used. This handles the aleatoric uncertainty with a negligible increase in computation, and reduces false positives, and improves the localization of bounding boxes.  Miller et al. \cite{miller2018dropout} applied dropout at test time to SSD~\cite{liu2016ssd} for improving uncertainty modeling capability. \cite{kraus2019uncertainty} applied MC dropout in YOLOv3 to model epistemic uncertainty and the Gaussian likelihood to model aleatoric uncertainty. \cite{harakeh2020bayesod} is also based on MC dropout to model epistemic uncertainty but used a Bayesian inference over per anchor bounding box. As dropout is ineffective for convolutional neural networks~\cite{ghiasi2018dropblock}, MC-Dropout-based approaches are also not effective in capturing uncertainty. Hence, we propose Monte Carlo DropBlock as an effective solution to improve generalization performance and uncertainty modeling capabilities in convolutional networks.

\section{Background}
We discuss here some basics of the Bayesian neural networks and deep learning models to understand their uncertainty modeling capabilities. We consider a supervised learning problem with inputs represented as $X=\{\bm{x}_{1}, \bm{x}_{2}, \ldots, \bm{x}_{n}\}$ where each $\bm{x}_{i}$ $\in$ $\mathbb{R}^d$ and corresponding outputs as $\textbf{y} = \{y_{1}, y_{2}.., y_{n}\}$ where each $ y_{i} \in \mathbb{R}$ for regression problems like bounding box detection and $y_{i} \in \{1,2,\ldots, C\}$ for classification problems. We aim to learn deep learning models which learn a mapping from  the input to output through linear-nonlinear transformations and convolution operations. 

\subsection{Bayesian Neural Networks}
\label{sec:bnn}
Bayesian neural networks (BNN) assume the  functional form of the model to be same as any other neural networks $f^{W}(\mathbf{x}) = W^L \sigma(W^{L-1} \ldots \sigma(W^1 \mathbf{x}))$ with activation function $\sigma$ and parameters $W=[W^{1}\ldots W^{K}]$. Let us assume the dimensionality of layer $l$ to be $K^l$, and consequently the weight matrix $W^l$ has dimension $K^l \times K^{l-1}$.  BNN assumes a distribution over the parameters and treats them as random variables. They assume a prior distribution over the weight vectors $p(W)$ which is often a Gaussian. The likelihood of generating output $y$ given $f^{W}(\mathbf{x})$ for some particular value of $W$ is modeled as a Gaussian in case of regression problems and Softmax in the case classification problems. BNNs estimate a posterior distribution over the parameters using the Bayes theorem. 
 \begin{equation}\label{eq:1}
     \begin{aligned}
         p(W \mid X,\textbf{y})=   \frac{\prod_{i=1}^N p(y_i\mid \bm{x}_i, W) \times p(W)}{p(\textbf{y}\mid X)}
     \end{aligned}
 \end{equation}
 
The prediction is done by considering multiple samples of weight vectors from the posterior and taking an average over them which is known as Bayesian model averaging. This will allow BNNs to capture uncertainty on predictions and make them resistant to overfitting. However,  posterior computation is intractable and requires using inference techniques such as Markov Chain Monte Carlo (MCMC) and Variational Inference (VI). For instance, VI assumes a particular distribution form for the posterior $q_{\theta}(W)= \prod_{i=1}^L q_{\theta^l}(W^l)$, where $q_{\theta}(W^l)$ is assumed to be a  Gaussian with parameters $\mu^l$ and $\Sigma^l$ (variational parameters $\theta^l$), and learns the variational parameters by minimizing the Kullback-Leibler (KL) divergence between $q(W)$ and $p(W \mid X,\textbf{y})$. As direct minimization of the KL divergence is intractable, VI learns the variational parameters $\theta$ by maximizing evidence lower bound (ELBO) which can be shown to equivalently minimize the KL divergence. The ELBO objective function in terms of  minimization can be written as follows
\begin{equation}\label{eq:2}
\begin{aligned}
L_{VI}(\theta)
=& -\sum_{i=1}^{n}\int q_{\theta}(W)\,log\,p(y_{i} \mid f^{W}(\bm{x_{i}}))\,dW  + &KL(q_{\theta}(W)\parallel p(W))
\end{aligned}
\end{equation}

\subsection{Monte Carlo Dropout}
Dropout\cite{JMLR:v15:srivastava14a} technique was introduced to avoid over-fitting in deep neural networks. During training, dropout randomly drops some of the neurons in every layer of a neural network based on the dropout probability $p$. This will prevent co-adaptation of neurons and mitigates over-fitting and lead to good generalization performance.  During testing, all the neurons are considered for prediction but each contributing according to the probability they are retained during the training phase. 


\cite{gal2016dropout} propose to apply dropout additionally at the testing time and showed that this will lead to a Bayesian neural network. The approach known as Monte-Carlo dropout (MC dropout) has become a very effective approach to obtain a Bayesian neural network capable of providing uncertainty estimates without the need to employ sophisticated inference techniques. 

Monte Carlo dropout can be shown to be equivalent to a Bayesian neural network employing variational\cite{lanczos2012variational} inference with a specific variational distribution. Dropout performed on the feature space can be seen as an operation in the weight space where random columns of the weight matrix mapping one layer to the next layer are dropped. This observation leads to the following form for the variational distribution $q(W^l)$ 
\begin{eqnarray}
\label{eq:vardistr}
& W^l = M^l \cdot diag([\bm{z}^l_j]_{j=1}^{K^{l-1}}) \\
& \bm{z}^l_j \sim Bernoulli(p)\, for \, j=1, \ldots,K^{l-1}  
\end{eqnarray}
where $M^l$ is the $K^l \times K^{l-1}$ dimensional weight vectors, and  $W^l$ follows a distribution defined by \eqref{eq:vardistr}. Consequently, $\hat{W}^l$ sampled from $q(W^l)$ will have some of the columns set to zero (leading to dropout) depending on $\hat{\bm{z}}^l_j$ sampled from $p(\bm{z}^l_j)$ which is $Bernoulli(p)$\footnote{Please note that we are considering the parameter $p$ of the Bernoulli distribution as probability of getting zero. }.
We can use this variational distribution in \eqref{eq:2} to derive the objective function as follows
\begin{eqnarray}\label{eq:do_obj}
& &  \hspace{-5mm} L_{DO}(M)
= -\sum_{i=1}^{n} \log\,p(y_{i} \mid f^{\hat{W}}(\bm{x_{i}})) + \lambda\parallel M \parallel^{2}  \\
& & \hspace{-5mm} \text{where} \quad \hat{W}=[\hat{W}^{1}...\hat{W}^{L}]=[diag(\hat{\bm{z}}^{1})M^{1}\ldots diag(\hat{\bm{z}}^{L})M^{L}] \nonumber 
\end{eqnarray}
We can easily observe that this objective function is the same as the one used to train the deep learning models with dropout regularization. The negative log-likelihood in  \eqref{eq:do_obj} is  cross-entropy loss for classification problems or least-squares loss for regression problems. We consider multiple $\hat{W}$ corresponding to different samples from $q(W)$ across different mini-batches during training.  BNN uses the posterior $q(W)$ for making predictions by sampling  $\hat{W}$ from $q(W)$ and computing an average over multiple samples of $\hat{W}$ (Monte Carlo sampling) and consequently over multiple dropout architectures.
\begin{equation}\label{eq:dropout_sampling}
\int q_{\theta}(W)\,p(y_{*} \mid f^{W}(\bm{x_{*}}))\,dW = \frac{1}{S} \sum_{s=1}^{S} \,p(y_{*} \mid f^{\hat{W}_s}(\bm{x_{*}}))
\end{equation}
Applying dropout at test time differentiates MC dropout from the standard dropout. The model averaging at test time allows it to model the uncertainty of the model in its predictions (epistemic uncertainty). Epistemic uncertainty estimation is especially useful when the model has to deal with out-of-distribution data. 


\subsection{Object Detection and Segmentation}


YOLO\cite{redmon2017yolo9000} is known for its one-stage detection and state-of-the-art performance. It's the most used network architecture in the world of object detection because of its fast speed and accuracy. In YOLO the last convolution layer splits the image into nxn cells, generally 19x19. Each of these cells is responsible for k bounding boxes (in general, k is chosen as 3). The training is faster because a single architecture performs both the tasks i.e. simultaneously predicting the multiple bounding boxes and class probabilities. In the convolutional layers of yolov3 (Figure \ref{fig:yolo}), kernels of shape 1x1 are applied on feature maps of three different sizes at three different places in the network. This leads to 3 predictions at 3 different scales, which happen due to downsampling the dimensions of the image by a stride of 32, 16, 8 respectively. Downsampling is done to reduce the size of data while maintaining the same spatial resolution. Every scale uses three anchor bounding boxes per layer. The three largest boxes for the first scale, three medium ones for the second scale, and the three smallest for the last scale. This way each layer excels in detecting large, medium, or small objects. The recent version of YOLO,  introduced by Alexey as YOLOv4\cite{bochkovskiy2020yolov4} and  YOLOv5\cite{glenn_jocher_2021_4679653}  are significant performance improvements compared to YOLOv3\cite{redmon2018yolov3}.

Instance Segmentation tasks have to generate masks, which are spatially coherent since the nearby pixels are more likely to belong to the same instance. A fully connected layer couldn't model the spatial features well compared to a fully convolutional layer. 
YOLACT (You Only Look At CoefficienTs)\cite{bolya2019yolact} has been experimented with because of its real-time inference capability with good accuracy for real-time instance segmentation. This method combined the qualities of single-stage methods and two-stage methods by parallelizing prototype mask generation and prediction of mask coefficients per instance. By choosing methods that are way above the typical real-time threshold (30 FPS), we have some flexibility of adding minor costs at inference without going to sub-realtime outputs.


\begin{figure}
\centering     
\subfigure[ Mask sampling in Dropout]{
\includegraphics[height =1.5in]{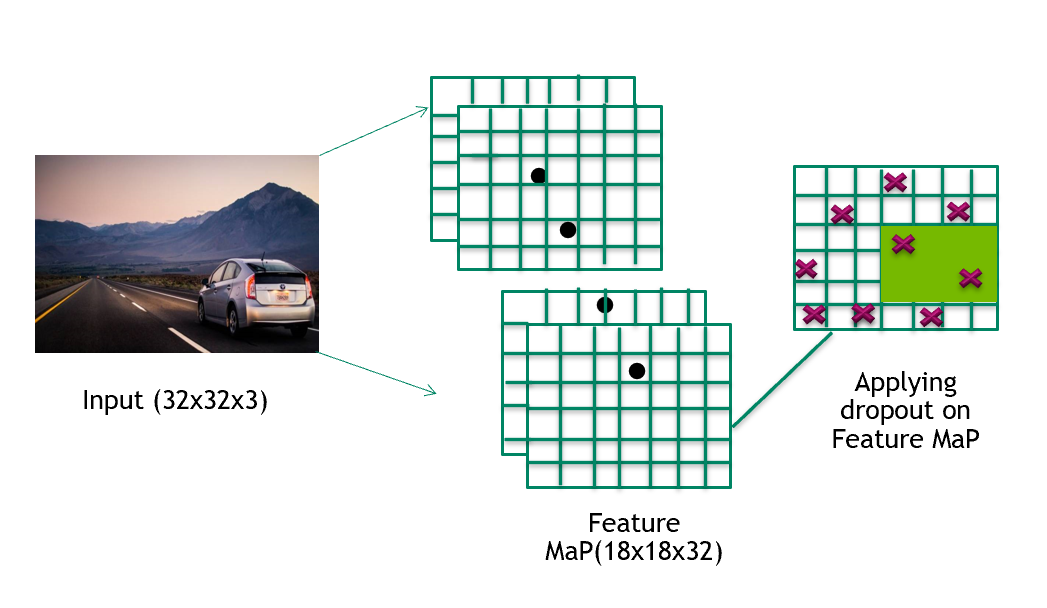}}
\qquad
\subfigure[Mask sampling in DropBlock ]{
\includegraphics[height =1.5in ]{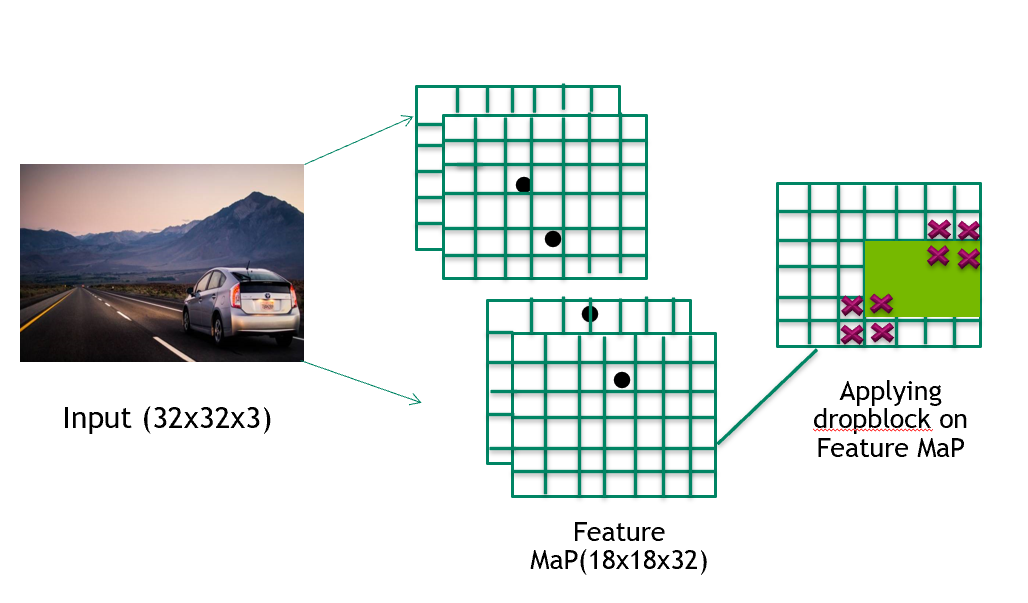}}
\qquad
\caption{(a) Mask sampling is completely random while in (b) during drop block masking is happening on the contiguous region as shown by the area marked by red.}
\label{fig:fig2}
\end{figure}

\section{Monte Carlo Drop Block}
The fully convolutional layers in YOLOv2\cite{redmon2017yolo9000} have resulted in significant improvement in object detection performance. Drop block~\cite{ghiasi2018dropblock} is quite useful to improve the generalization performance in the deep learning models involving convolutional. layers such as Resnet and Imagenet.  When dropout was added to convolutional layers,  it is found to be ineffective as compared to its use in fully connected layers.  DropBlock drops blocks of some particular size with some drop probability in the convolutional layer. Dropping contiguous regions from the feature map results in better regularization of neurons and improvement in generalization performance. Figure~\ref{fig:fig2} provides differences in Dropout and DropBlock based sampling of neuron masks and subsequent feature representation.

We propose the Monte-Carlo DropBlock inference technique for deep learning models, which could provide the desired uncertainty modeling capabilities in solving complex vision tasks. In particular, the proposed approach models the epistemic uncertainty by applying drop block at the training stage and the inference stage (Algorithm~\ref{alg:algorithm}). 
MC Dropout was shown to improve the epistemic uncertainty modeling capability of fully connected neural networks~\cite{kendall2017uncertainties}.  Using DropBlock at inference time can be seen as a similar approach to MC Dropout which does multiple predictions using a different architecture obtained through dropout. Applying DropBlock during inference can be interpreted as an averaged prediction across an exponential ensemble of sub-networks. This can be seen as equivalent to a Bayesian convolutional neural network with a more structured variational posterior for inference as we show in the following section.

\begin{algorithm}[tb]
\caption{Monte Carlo DropBlock}
\label{alg:algorithm}
\textbf{Input}: output from layer (X)\\
\textbf{Parameters}: mode, \textit{block\_size}, $\gamma $\\
\begin{algorithmic}[1] 
\IF {( mode == training or mode == inference)}
\STATE Randomly sample mask \textit{M}: \textit{$M_{i,j}$} $\sim$ \textit{Bernoulli($\gamma$)}
\STATE For each position having zero \textit{$M_{i,j}$} , create a contiguous square mask with \textit{$M_{i,j}$} as the center, $block\_size$ as width and height. Then set all the values of M in the square to be zero
\STATE Apply the mask: X = X $\times$ M 
\STATE Normalize the features: X = X $\times$ count(M) = $count\_ones(M)$ 
\ENDIF 
\end{algorithmic}
\end{algorithm}

\subsection{Drop block as Bayesian Approximation}

In this section, we show that the drop block applied to the feature maps in the neural network layers at train time and test time will result in a Bayesian neural network. As already discussed in Section \ref{sec:bnn}, approximate inference techniques for BNNs such as variational inference assumes a specific form for the posterior to be learned.  We show that a particular form of the variational posterior will result in operations similar to applying drop block at training time and test time in deep learning models.

Let us assume the feature representation obtained at some particular layer $l$ to be represented as $A^l$. For ease of understanding, let's assume it's a square matrix of size $K^l \times K^l$~\footnote{The discussion can be generalized to feature matrix, filter matrix, block size, and stride lengths of arbitrary size and to multiple channels.}. Consider a convolution operation using a filter (kernel) $W^l$ of size $L \times L$. The feature representation at the next layer is obtained by performing a convolution operation over the feature matrix $A^l$ and filter $W^l$ assuming a stride length $L$. This will lead to a feature representation $A^{l+1}$ of size $\frac{K^l}{L} \times \frac{K^l}{L}$. There are ${\frac{K^l}{L}}^2$ blocks in the feature matrix $A^l$, each of them are convolved against the filter $W^l$ to obtain the $\frac{K^l}{L}^2$ feature vector. Now let us consider applying drop blocks in the layer $l$ with drop probability $\gamma$ and block size $L \times L$. Assume $i^{th}$ block of size $L \times L$ in the feature matrix $A^l$ got masked as a result of applying drop block. Consequently, the $i^{th}$ feature representation in layer $l+1$, $A^{l+1}(i)$ will be zero. 
\begin{figure}[!tbp]
    \centering
    \includegraphics[width = 4.5in]{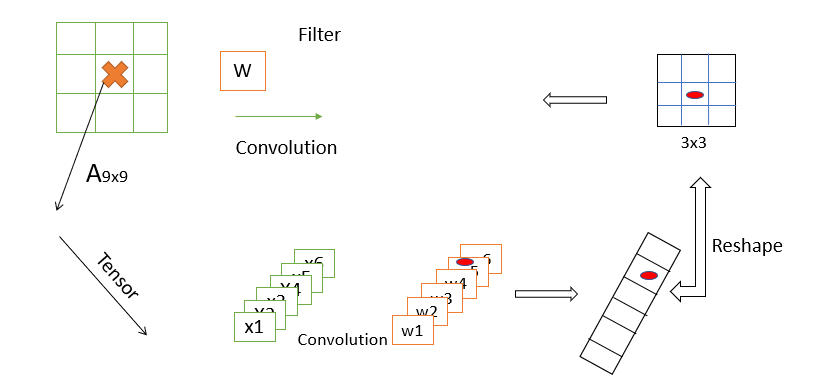}\\
    \caption{Variational Drop block: Representing dropBlock as operations over 3-dimensional tensor representation of feature matrix and weight vectors. }
    \label{fig:my_label}
\end{figure}

To apply MC dropout theory, we need to first represent drop block operations as operations using the weight vectors (filters in the convolution operation). We can observe that this can be achieved by considering a larger weight matrix $\bar{W}^l$ of size $K^l \times K^l$ with $\frac{K^l}{L}^2$ blocks, with each block being the $L \times L$ matrix $W^l$, except the $i^{th}$ block being all zeros. For brevity, we consider the feature matrix $A$ as a 3-dimensional tensor with $L \times L$ blocks stacked up in the third dimension of size $\frac{K^l}{L}^2$. Similarly, $\bar{W}^l$ is a 3-dimensional tensor with $W^l$ blocks stacked one after the other and a block with zeros as the $i^{th}$ element in the third dimension. We can now think of getting the feature vector $A^{l+1}$ as performing convolution between feature matrix $A^l$ and weight vector $\bar{W}^l$ element-wise in the third dimension (see Figure~\ref{fig:my_label}). 

We can see that considering  $\bar{W}^l$ with blocks of all zeros appearing as some element in the third dimension of the tensor is equivalent to considering a feature representation with some masked block.  Now we see how to define a variational distribution over $\bar{W}^l$ which will lead to a sample of $\bar{W}^l$ as we considered. To define the variational distribution, we consider the tensor $\bar{W}$ to consist of blocks $W_1, W_2, \ldots, W_{\frac{K^2}{L^2}}$ stacked one after the other in the third dimension. We assume the variational distribution $q(\bar{W})$ can be decomposed as the product of the variational distributions over $W_i$'s, i.e. $q(\bar{W}) = \prod_i q(W_i)$. The blocks $W_i$ can take value $0$ with drop probability $\gamma$ or can take value $W$ with probability $1-\gamma$. Thus, we define the variational distribution $q(W_i)$ over the block $W_i$ as  $W_i = W \cdot z_i$ and $z_i$ sampled from $Bernoulli(\gamma)$\footnote{Dropblock paper considered the probability of getting $0$ as the parameter to the Bernoulli distribution and we follow the same convention.}. Thus the variational distribution over $\bar{W}$ can be summarized as
\begin{eqnarray}
& & q(\bar{W}) = \prod_i q(W_i) \nonumber  \\
& & q(W_i) = W \cdot z_i   \qquad 
z_i \sim Bernoulli(\gamma)
\end{eqnarray}

\subsection{Modeling Aleatoric uncertainty }

To predict the aleatoric uncertainty in bounding boxes, Gaussian Yolo~\cite{choi2019gaussian} proposes to model each of the four coordinates as a Gaussian (with mean and variance parameters ) and re-designing the loss function as negative log-likelihood (NLL). As a result, the number of outputs for the bounding box becomes 8($\mu$ and $\sigma$ for four coordinates) instead of 4 (x,y,w,h). This will lead to a change in detection criteria by including localization uncertainty. We consider a Generalized IoU(GIoU)\cite{rezatofighi2019generalized} loss, which uses orientation and shape of the object in addition to the area covered. This gives a better result when used along with Gaussian YOLO for object detection. We combine the MC-DropBlock approach with the Gaussian Yolo approach, resulting in object detection models which are capable of modeling both epistemic and aleatoric uncertainties. 


\section{Experiments}

In this section, we experimentally evaluate the performance of the proposed MC DropBlock approach to model uncertainty in YOLO models for object detection and segmentation. Additionally, we also demonstrate the ability of MC-DropBlock to model uncertainty in image classification models such as ResNets.   We consider different ways to apply DropBlock in YOLO models and show how it can improve the uncertainty modeling, generalization performance, and calibration capabilities of the YOLO models. We compare the proposed  MC DropBlock with various baselines including the ones considering drop block only at training time (\emph{training time dropBlock}) or only at inference or test time (\emph{inference time dropBlock}), and with existing approaches such as Dropout based \cite{azevedo2020stochastic} and Gaussian Yolo models \cite{feng2020review}.  All the experiments are trained with NVIDIA V100 GPUs.

\begin{figure}
\centering

\includegraphics[width = 4.5in]{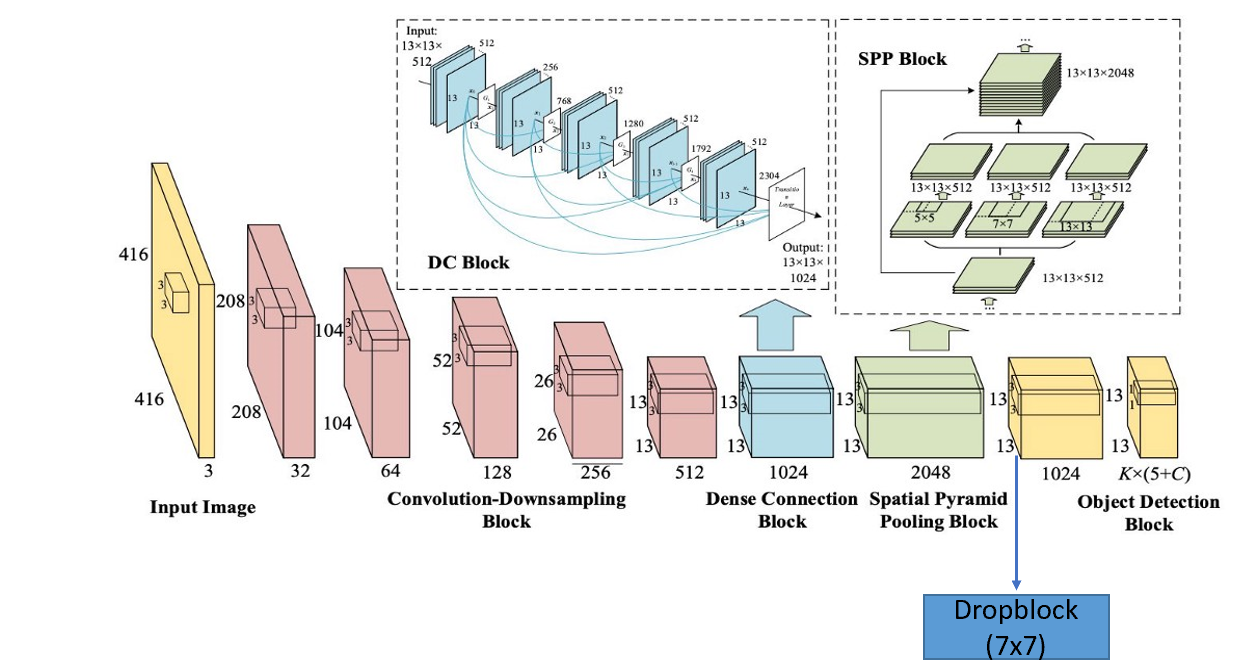}

\caption{YOLOv4 architecture where drop block is implemented after spatial pyramid pooling block} 
\label{fig:yolo}
\end{figure}

We applied DropBlock at the later layers of the YOLO and  ResNet models as initial layers capture the basic and important features from
the image while the later layers extract very high-level features. Our experiments also revealed that applying DropBlock at initial layers leads to a significant
drop in performance. For instance, in YOLOv5 architecture, DropBlock is applied before the Detect module in the prediction head. In YOLOv4 architecture, we used CSPDark-net53 as the backbone. DropBlock is applied after the Spatial Pyramid Pooling(SPP)  block as shown in Figure \ref{fig:yolo} and before the prediction head.  For Instance, segmentation using YOLACT, DropBlock is applied before the prediction head in the feature pyramid. We used Resnet-50 + FPN backbone because of its high speed compared to the bigger and slower Resnet-101. Similarly, for classification with Renet-110, DropBlock was applied before the prediction head.

The proposed MC drop block approach aims to enhance YOLO models with uncertainty modeling capability so that they do not make overconfident predictions when presented with a test data point from a distribution different from training data. Towards this, we conduct the out-of-distribution (OOD) experiments, by training on the Pascal VOC~\cite{everingham2010pascal} dataset and testing on the COCO dataset.
We also study the generalization performance of the proposed approach on the COCO dataset~\cite{lin2014microsoft} for object detection and segmentation experiments, and on CIFAR-10 for image classification experiments. For capturing generalization performance, we used the standard evaluation metrics such as mean Average Precision (mAP)~\cite{redmon2016you} for object detection and object segmentation, and accuracy for image classification.
In addition, We considered two evaluation metrics, the Brier score to check the calibration capability of the model and the entropy score for measuring the ability to model uncertainty on OOD experiments~\cite{jain2020decision}, which we describe in detail in the following paragraph.   Entropy is the established measure of capturing uncertainty~\cite{gal16} and is used in several Bayesian deep learning models~\cite{malinin2018predictive,hullermeier2021aleatoric}. Through these experiments, we aim to show that  MC-DropBlock brings significant improvement in uncertainty modeling capability while maintaining good generalization and calibration capabilities.

\textbf{Brier Score : } Brier score measures the calibration capability of a model~\cite{gail85}, i.e. if the predictive probability matches the true underlying distribution of labels in the data set.  It is calculated as the squared error between a predictive probability associated with class labels and the one-hot encoded ground-truth label in the test data set.  
Given $N_*$ number of test samples, for a classification problem with \textit{C} classes, Brier Score (BS) is obtained by:
\begin{equation}
    BS = \frac{1}{N_*} \sum_{n=1}^{N_*} \sum_{c=1}^ C \bigl (p(y_{*n} = c|\bm{x}_{*n}, D) - y_{*n,c} \bigr )^2
\end{equation}
where $y_{*n,c}$  is one hot vector encoding of ground truth label (“1” if the ground truth class is c and “0” otherwise).
Brier scores range between 0 and 1, with smaller values indicating better calibration capability.

\textbf{Entropy : } We compute the Shannon-entropy which is the expectation of the negative log of the conditional probabilities that the test input $\bm{x}_*$ is assigned a label $y_*$.
\begin{equation}
    H(\bm{x}_*, D) = - \sum_{c=1}^C p(y_* = c|\bm{x}_*, D) \log p(y_*=c|\bm{x}_*, D)  \nonumber 
\end{equation}
Entropy is higher when classes are equiprobable and low otherwise. On OOD data, we expect the model to be highly uncertain about its predictions and not make high confident (probability) predictions. Consequently, we favor models with high values of entropy on OOD data. 



\subsection{Out-of-Distribution Experiments}
For out-of-distribution (OOD) experiments for object detection, we trained our model on the Pascal VOC dataset and then tested it on the COCO dataset.    Pascal VOC \cite{everingham2010pascal} data has 11,530 images belonging to 20 classes. COCO dataset~\cite{lin2014microsoft} for the object detection task consisting of approximately 164K images across 91 classes. All the 20 classes of the Pascal VOC dataset (car, person, bus, etc) are also present in the COCO dataset. COCO data set contain classes, not in Pascal VOC and hence we expect the model to exhibit high uncertainty on COCO in this experimental setup. The hyperparameters used in YOLO models are the same as described in the object detection task.
\begin{table*}[t]
  \centering
  \resizebox{\linewidth}{!}{%
   \begin{tabular}{ |c|c|c|c|c|c|c|c|}
\hline

 \text{Model} & \text{Training data} &\textbf{Testing data} & \textbf{with out DropBlock}  & \textbf{Training time DropBlock}& \textbf{MC-Dropout} & \textbf{MC Dropblock} \\
 \hline
         YOLOv5  & Pascal VOC & COCO  &  0.3147 & 0.3038 & 0.35 & \textbf{0.3895} \\
         GaussianYOLOv5  & Pascal VOC & COCO  &  0.3148 & 0.3222 & 0.3514 & \textbf{0.3546} \\
         YOLOv4  & Pascal VOC & COCO & 0.4023 & 0.4145 & 0.4328 & \textbf{0.4526}\\
         Gaussian YOLOv4  & Pascal VOC & COCO &  0.3967 & 0.4056 & 0.4278 & \textbf{0.4434}\\
         ResNet  &   CIFAR-10 & ImageNet & 0.2675 & 0.2788 & 0.2821 & \textbf{0.2901}\\
        
\hline

\end{tabular}}
  \caption{Entropy values of different models for out-of-distribution experiments}
  \label{tab:5}
\end{table*}
We have used entropy as the evaluation metric for measuring uncertainty. Table \ref{tab:5} shows the results of  out-of-distribution experiments. We can see that  MC-DropBlock provided the highest entropy for various YOLO models as well as ResNets. It outperformed  MC-Dropout and other ways of adding DropBlock in these models for object detection and classification tasks. 

\subsection{Object Detection task}
We have used the same experimental setup as \cite{bochkovskiy2020yolov4} for our experiments on YOLOv4 with COCO 2017 dataset having training/validation split as 118K/5K and the test split has 41K. The hyperparameters for YOLOv4 and Gaussian YOLOv4 experiments are: 500500 training steps, learning rate 0.01 multiplied with a factor of 0.1 at 400000 steps and 450000 steps, momentum of 0.9, weight decay of 0.0005, batch size of 64, and mini-batch size of 4. For our experiments on  YOLOv5, we have used the same setup as \cite{glenn_jocher_2021_4679653}. YOLOv5 has 4 variations(s, m, l, x) for various applications. We experiment on the YOLOv5(x) model is the same as YOLOv4 and gives state-of-the-art mAP. The experimental setup for YOLOv5 and Gaussian YOLOv5 is the same. We have  experimented with various drop block sizes ($3 \times 3,5 \times 5,7 \times 7$) and drop probabilities ($0.1, 0.3,0.5,0.7$). We found that $7 \times 7$ drop block size with drop probability $0.1$ gave optimal mAP score (on validation data) for MC-DropBlock and \emph{training time dropBlock},   While for \emph{inference time dropBlock} $3 \times 3$ gave the best  result. 

\begin{table*}[t]
  \centering
  \resizebox{\linewidth}{!}{%
  \begin{tabular}{ |c|c|c|c|c|c|} 
\hline

Model &\textbf{IoU}  & \text{YOLOv4}&\textbf{YOLOv5x } & \textbf{Gaussian YOLOv4} & \textbf{Gaussian YOLOv5x} \\
\hline
without DropBlock & 0.5  & 0.6434 &  0.69  & 0.6627 & 0.7108\\
Training time DropBlock & 0.5 & \textbf{0.6523} & \textbf{0.7050}& \textbf{0.6718}& \textbf{0.7261} \\
inference time DropBlock & 0.5 &  0.6515 & 0.6995 & 0.6710& 0.7204 \\
MC-DropBlock & 0.5 &  0.6478 	& 0.7023&	 0.6714 & 0.7233\\
MC Dropout & 0.5 &  0.628 	& 0.669& 0.64 & 0.676 \\
\hline
\end{tabular}
}
\\
  \caption{mAP values for Object detection on the COCO dataset}
  \label{tab:1}
\end{table*}
Table \ref{tab:1} shows the mAP score of the models trained and tested on the COCO dataset. A higher mAP value indicates better precision. From the Table, we can observe that the YOLOv4/v5 models give the best mAP value for \emph{train time DropBlock}, and MC-DropBlock gives the second-best mAP value.  As we see later, MC-DropBlock in addition provides a substantial improvement in the uncertainty modeling capabilities of these models.   We can also observe that adding  Gaussian loss to model aleatoric uncertainty has improved the mAP score of the YOLO models.

%

%




Table \ref{tab:3} provides the calibration ability of the models measured in terms of the Brier score.   We can see from Table \ref{tab:3} Brier score is the lowest for both the algorithm for the \emph{Inference time DropBlock}. In the calibration experiments, MC-DropBlock gives the second-best result and is better than the one without DropBlock or with training time drop block alone. 
 \begin{table*}[t]
  \centering
  \resizebox{\linewidth}{!}{%
  \begin{tabular}{ |c|c|c|c|c|c|} 
\hline

Model &\textbf{IoU}  & \text{YOLOv4}&\textbf{YOLOv5x } & \textbf{Gaussian YOLOv4} & \textbf{Gaussian YOLOv5x} \\
\hline
without DropBlock & 0.5  & 0.4298 &  0.3523  & 0.4169 & 0.3417\\
Training time DropBlock & 0.5 & 0.4199 & 0.3371 & 0.4073 & 0.3269 \\
inference time DropBlock & 0.5 &  \textbf{0.4175} & \textbf{0.3328} & \textbf{0.4049} & \textbf{0.3228} \\
MC-DropBlock & 0.5 &  0.4187 	& 0.3349 &	 0.4061 & 0.3248 \\
\hline
\end{tabular}
  }
\\
  \caption{Brier score values of different models for Object detection on the COCO dataset.}
  \label{tab:3}
\end{table*}






\subsection{Segmentation task}
For instance segmentation, we conducted experiments on the MS COCO~\cite{lin2014microsoft} segmentation data. It has pixel-wise mask value along with bounding box value in annotation files. For MS COCO, we train on train2017 and evaluate on
val2017 and test-dev. We conducted experiments with the same setup as   YOLACT-550\cite{bolya2019yolact} segmentation model on COCO 2017 dataset. 
We performed our experiments with batch size 8 on a single GPU using pre-trained weights of ImageNet~\cite{deng2009imagenet} and with the same configuration used by the authors. For YOLACT-550 with Resnet-50-FPN, we experimented with DropBlock having block size 7x7 and drop probability of 0.1 before the prediction head.  From Table \ref{tab:2} we can see \emph{inference time DropBlock} shows an improvement of \textbf{1.1\%} on mask mAP value, which measures the quality of segmentation mask and \textbf{1.05\%} improvement on box mAP, which measures the quality of bounding boxes. MC-Drop block and \emph{training time DropBlock} also give a better mAP value than the baseline model.  
\begin{table*}[t]
  \centering
  \resizebox{\linewidth}{!}{%
  \begin{tabular}{ |c|c|c|c|c|c|} 
\hline

Model &\textbf{IoU} & \text{Yolact550 (mask mAP)}&\textbf{Yolact550 (box mAP) } \\
\hline

without DropBlock & 0.5 & 0.282 & 0.303 \\

Training time DropBlock & 0.5 & 0.29 & 0.3124 
\\
Inference time DropBlock& 0.5 & \textbf{0.293} & \textbf{0.3135}
\\
MC-DropBlock& 0.5 & 0.2917 & 0.3122 
\\
\hline
\end{tabular}}

  \caption{mAP value on COCO dataset from YOLACT550 segmentation model }
  \label{tab:2}
\end{table*}

\subsection{Classification task}
We also experimented with ResNet-110~\cite{vijayakumar2019classification} model for image classification on the CIFAR-10 Dataset \cite{krizhevsky2009learning}. CIFAR-10 contain images from 10 classes with 50000 images for training  and 10000 images for testing. DropBlock with block size $7 \times 7$ and drop probability $0.1$ is added after the final convolutional layer for MC DropBlock. For MC Dropout, drop probability is set to $0.5$. We obtained $92.08\%$ top-1 accuracy using baseline ResNet-110 model, $92.26\%$ top-1 accuracy using MC Dropout and \textbf{92.98\%} top-1 accuracy using MC DropBlock. 
We conducted out-of-the-distribution experiments for image classification by training on CIFAR-10 and testing on the  Imagenet~\cite{russakovsky2015imagenet}. Imagenet consists of 1000 classes and the experiments are conducted on a test set consisting of 1,50,000 images.   As observed in Table~\ref{tab:5}, we obtained the highest entropy score for the MC-Drop block, and consequently, it has the best uncertainty modeling capability on image classification models like Resnet-110.

\begin{figure*}[!ht]%
\centering
\subfigure[Baseline YOLOv5 ]{%
\label{fig:first}%
\includegraphics[width = 1.25in, height=1in]{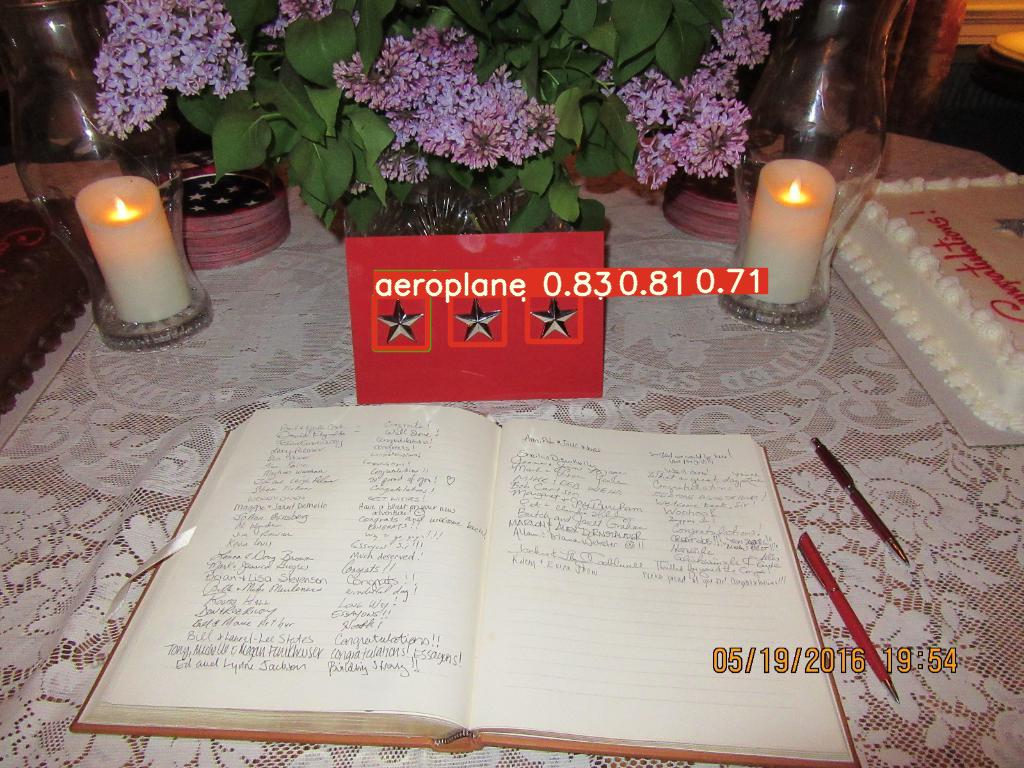}}%
\qquad
\subfigure[MC-Dropout YOLOv5 ]{%
\label{fig:second}%
\includegraphics[width = 1.25in,height=1in]{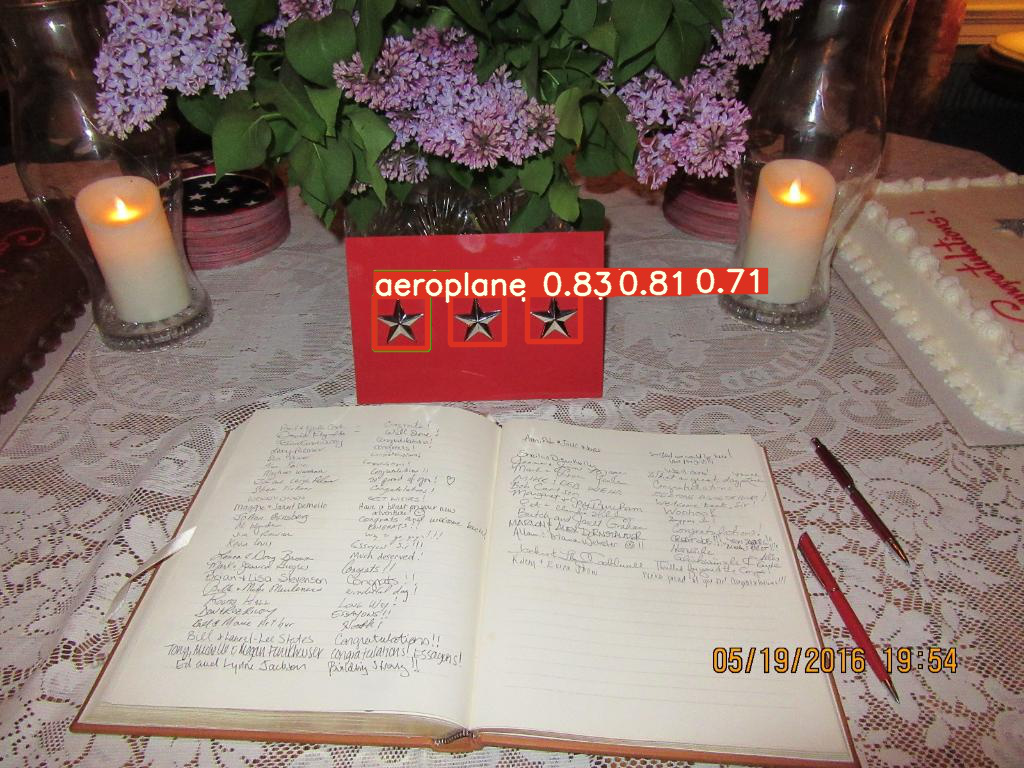}}%
\qquad
\subfigure[MC-DropBlock YOLOv5 ]{%
\label{fig:third}%
\includegraphics[width = 1.25in,height=1in]{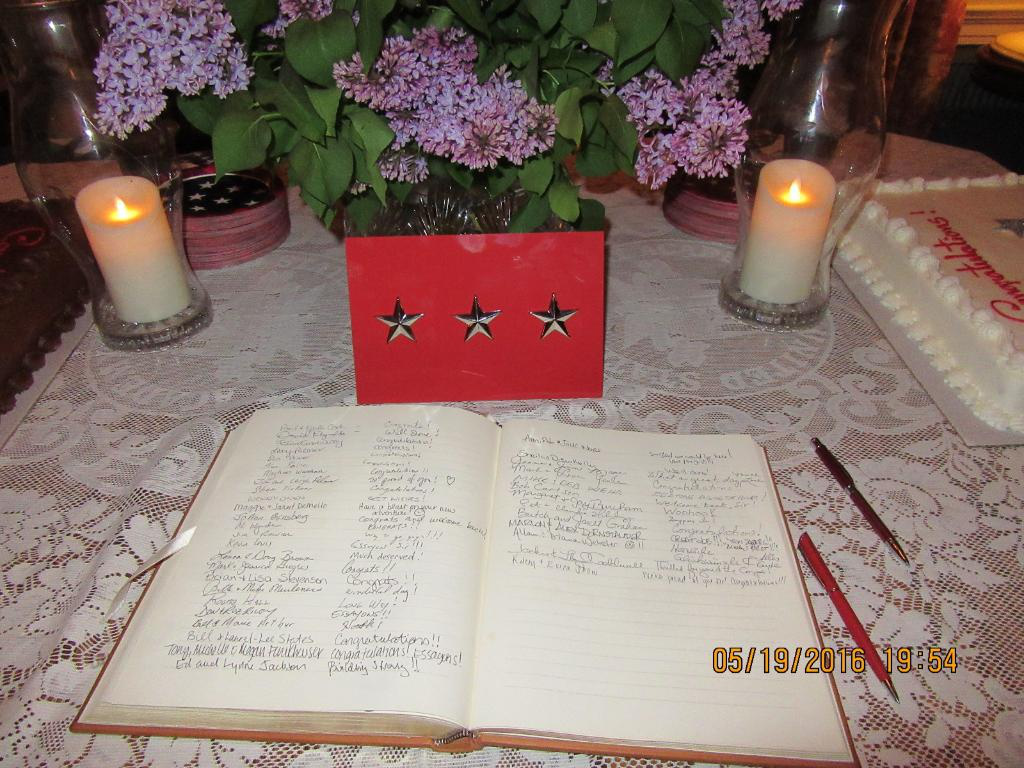}}%
\caption{Results of YOLOv5 trained on COCO and tested on an image from Open Images Data. Baseline model and MC-Dropout are miss-classifying pot design as 3 different aeroplane objects. MC-DropBlock completely removed the miss-classification. }
\label{fig:sample1}
\end{figure*}

\begin{figure*}[!ht]%
\centering
\subfigure[YOLOv5 Baseline]{%
\label{fig:first2}%
\includegraphics[width = 1.25in, height=1in]{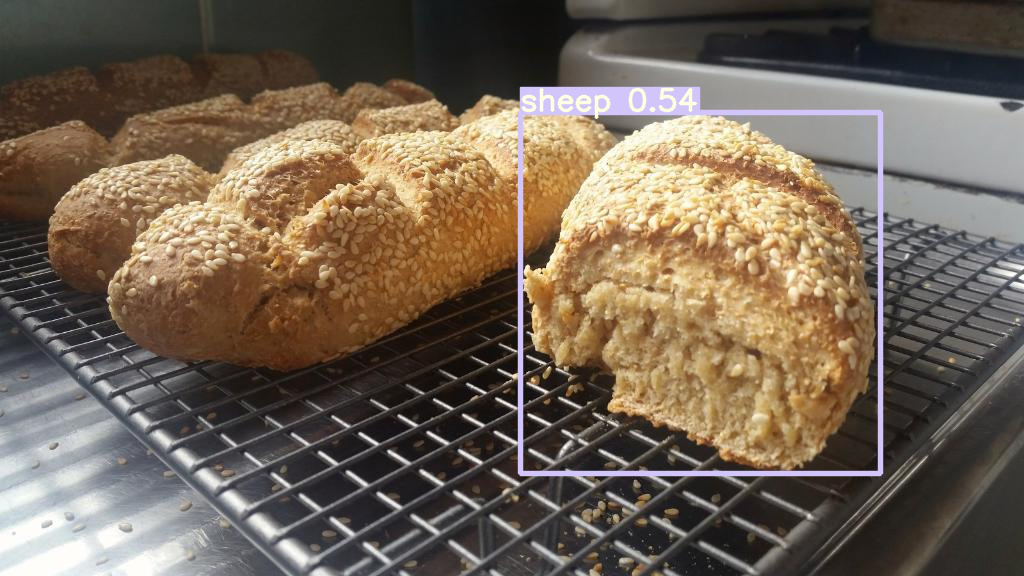}}%
\qquad
\subfigure[YOLOv5 MC-Dropout]{%
\label{fig:second2}%
\includegraphics[width = 1.25in, height=1in]{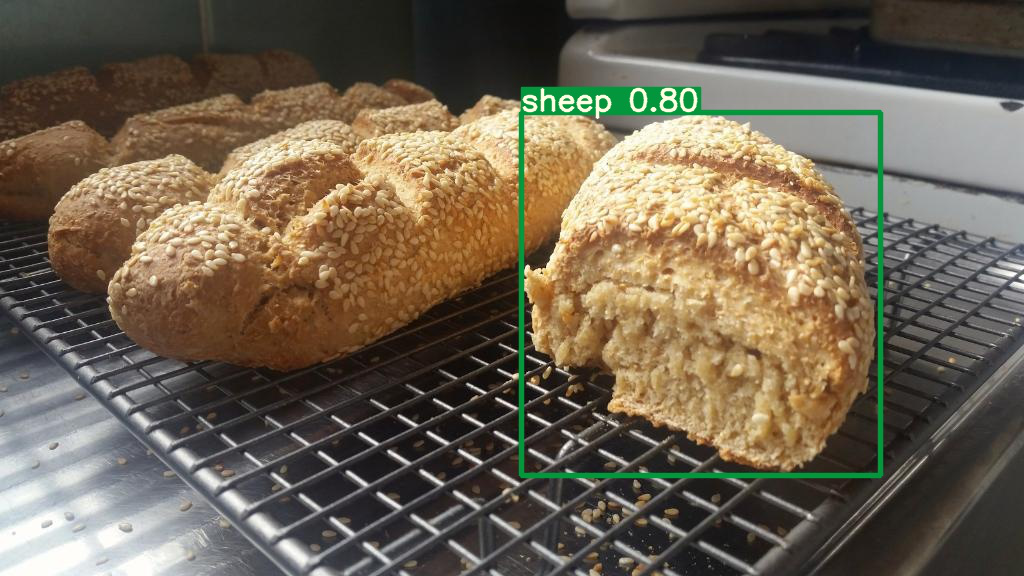}}%
\qquad
\subfigure[YOLOv5 MC-DropBlock]{%
\label{fig:third2}%
\includegraphics[width = 1.25in, height=1in]{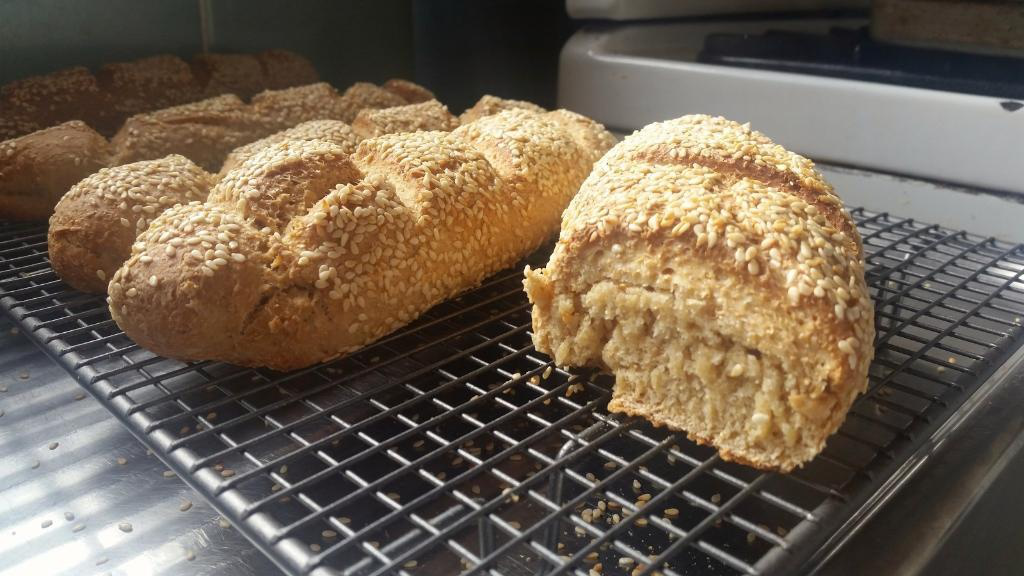}}
\caption{Results of YOLOv5 trained on COCO and tested on an image from Open Images Data. The baseline model and MC-Dropout are miss-classifying bread as sheep. MC-DropBlock completely removed the miss-classification. }
\label{fig:sample2}
\end{figure*}

\subsection{Qualitative Analysis}
We conduct qualitative analysis to study the out-of-distribution object detection capabilities of the proposed MC-DropBlock approach on YOLOv5.  We train our models on  Pascal VOC and tested on images from Open Images Dataset~\cite{OpenImages2} and COCO. From Figure~\ref{fig:sample1}, we can observe that the  baseline YOLOv5 model (Figure~ \ref{fig:first}) and  MC-Dropout on YOLOv5 (Figure~ \ref{fig:second})  identifies aeroplane wrongly.  While the  MC-DropBlock (Figure~ \ref{fig:third}) on YOLOv5, no longer miss-classifies the object as an aeroplane. Similarly, in Figure~\ref{fig:sample2}, baseline YOLOv5 and MC-Dropout on YOLOv5 detected the bread loaf incorrectly as sheep, while MC-DropBlock does not make any false predictions. 

\section{Conclusion}
In this work, we have developed an effective approach based on Monte Carlo DropBlock to model uncertainty in complex convolutional neural networks for vision problems. We have shown that  MC-DropBlock can be seen as performing variational inference on a  Bayesian convolutional neural network. Incorporating DropBlock at test time proved to be an effective approach for improving calibration and uncertainty modeling capability along with improving generalization performance.   Further Gaussian loss is incorporated in YOLO models to handle aleatoric uncertainty as well. MC DropBlock is a better method for fully convolutional layer-based architectures than MC Dropout for modeling uncertainty. The proposed method has been tested on multiple vision tasks and the experimental results show that any vision task which uses a fully convolutional layer could benefit out of MC-DropBlock.   
\section{Acknowledgments}
We thank the   support from Nvidia, India and Nvidia AI Technology Center (NVAITC), IIT Hyderabad for providing the computing power and compute infrastructure requirements for the project. 

\bibliographystyle{IEEEtran}
\bibliography{IEEEabrv, egbib}

\end{document}